\documentclass[12pt,a4paper]{article}

\usepackage{geometry}
\usepackage{graphicx,url}
\usepackage{eurosym}

\usepackage{cite}
\usepackage[cmex10]{amsmath}
\usepackage{algorithmic}
\usepackage{algorithm}

\usepackage{array}
\usepackage{mdwmath}
\usepackage{mdwtab}
\usepackage{eqparbox}

\usepackage{fixltx2e}
\usepackage{url}
\usepackage{float}
\providecommand{\keywords}[1]{\textbf{\textit{Keywords:}} #1}

\usepackage[toc,nonumberlist]{glossaries}
\glsdisablehyper
\newacronym{uav}{UAV}{Unmanned Aerial Vehicle}
\newacronym{cep}{CEP}{Complex Event Processing}
\newacronym{faa}{FAA}{Federal Aviation Administration}
\newacronym{epl}{EPL}{Event Processing Language}
\newacronym{sql}{SQL}{Structured Query Language}
\newacronym{pojo}{POJO}{Plain Old Java Object}

\newacronym{u2ud}{U2UD}{UAV-to-UAV Detections}
\newacronym{sod}{SOD}{Static Obstacle Detections}
\newacronym{mod}{MOD}{Moving Obstacle Detections}

\newacronym{u2uc}{U2UC}{UAV-to-UAV Collisions}
\newacronym{soc}{SOC}{Static Obstacle Collisions}
\newacronym{moc}{MOC}{Moving Obstcale Collisions}

\newacronym{ahc}{AHC}{Average Hovering Count}
\newacronym{abc}{ABC}{Average Backtracking Count}
\newacronym{cat}{CAT}{Collision Avoidance Time}

\newacronym{pso}{PSO}{Particle Swarm Optimization}
\newacronym{ga}{GA}{Genetic Algorithm}

\newacronym{sr}{SR}{Safety Requirement}

\newacronym{rrt}{RRT}{Rapidly-Exploring Random Trees}

\newacronym{arl}{ARL}{Average Route Length}
\newacronym{llr}{LLR}{Length of the Longest Route}
\newacronym{nc}{NC}{Number of Collisions}
\newacronym{t}{T}{Computation Time}

\makeglossaries

\usepackage{url}

\usepackage{listings} 

\usepackage{algorithmic}
\usepackage{algorithm}

\usepackage{graphicx} 
\usepackage{subfig} 

\usepackage{multirow} 

\begin{document}

\date{}

\title{{\large\textbf{Online Path Generation and Navigation for Swarms of UAVs}}}

\author{
{\small Adnan Ashraf$^{\ast}$, Amin Majd$^{\ast}$, Elena Troubitsyna$^{\ast\dagger}$}\\
{\footnotesize $^{\ast}$Faculty of Natural Sciences and Technology, \AA bo Akademi University, Finland} \\ 
{\footnotesize \texttt{adnan.ashraf@abo.fi}, \texttt{amin.majd@abo.fi}, \texttt{elena.troubitsyna@abo.fi}} \\
{\footnotesize $^{\dagger}$Department of Theoretical Computer Science, KTH Royal Institute of Technology, Sweden} \\ 
{\footnotesize \texttt{elenatro@kth.se}}
}

\maketitle

\begin{abstract}
With the growing popularity of \glspl{uav} for consumer applications, the number of accidents involving \glspl{uav} is also increasing rapidly. Therefore, motion safety of \glspl{uav} has become a prime concern for \gls{uav} operators. For a swarm of \glspl{uav}, a safe operation can not be guaranteed without preventing the \glspl{uav} from colliding with one another and with static and dynamically appearing, moving obstacles in the flying zone. In this paper, we present an online, collision-free path generation and navigation system for swarms of \glspl{uav}. The proposed system uses geographical locations of the \glspl{uav} and of the successfully detected, static and moving obstacles to predict and avoid: (1) \gls{uav}-to-\gls{uav} collisions, (2) \gls{uav}-to-static-obstacle collisions, and (3) \gls{uav}-to-moving-obstacle collisions. Our collision prediction approach leverages efficient runtime monitoring and \gls{cep} to make timely predictions. A distinctive feature of the proposed system is its ability to foresee potential collisions and proactively find best ways to avoid predicted collisions in order to ensure safety of the entire swarm. We also present a simulation-based implementation of the proposed system along with an experimental evaluation involving a series of experiments and compare our results with the results of four existing approaches. The results show that the proposed system successfully predicts and avoids all three kinds of collisions in an online manner. Moreover, it generates safe and efficient \gls{uav} routes, efficiently scales to large-sized problem instances, and is suitable for cluttered flying zones and for scenarios involving high risks of \gls{uav} collisions.
\end{abstract}


\glsresetall

\section{Introduction}

An \gls{uav} or drone is a semi-autonomous aircraft that can be controlled and operated remotely by using a computer along with a radio-link~\cite{Austin:2011}. \Glspl{uav} can be classified into different types based on their design, size, and flying mechanism. Among the existing types, the quadrotors or quadrocopters are particularly popular because of their simple design, small size, low cost, greater maneuverability, and the ability to hover-in-place. A quadrotor uses two pairs of identical, vertically oriented propellers of which one pair spins clockwise and the other spins counterclockwise. Commercially-available quadrotors are increasingly been used in a variety of applications such as monitoring and surveillance, search and rescue operations, geographic mapping, photography and filming, wildlife research and management, media coverage of public events, remote sensing for agricultural applications, and aerial package delivery~\cite{Ashraf:2017:ECBS, Dong:2009, Ivanovas:2018, Pajares:2015}. Efficient and scalable solutions for these applications require an online path generation and navigation system for multiple \glspl{uav}.

\Glspl{uav} are becoming increasing popular. In the United States, the \gls{faa} has projected that the number of small hobbyist drones is set to increase from an estimated 1.1 million in 2017 to 2.4 million by 2022\footnote{\url{https://www.faa.gov/news/updates/?newsId=89870}}. With the growing popularity and use of \glspl{uav} for consumer applications, the number of accidents involving drones is also increasing dramatically. The \gls{faa} receives more than 100 reports every month of unauthorized and potentially hazardous \gls{uav} activity reported by pilots, citizens, and law enforcement\footnote{\url{https://www.faa.gov/uas/resources/public_records/uas_sightings_report/}}. In a recent incident\footnote{\url{https://www.bbc.com/news/uk-england-sussex-46623754}} that took place in the United Kingdom, the runway at the London Gatwick Airport was shutdown for more than a day because two drones were spotted flying repeatedly over the airfield. The disruption affected about 110,000 passengers on 760 flights as no flights were able to take off or land. Such incidents on one hand show the importance of educational and training programmes for drone operators and stricter legislation for offenders, but on the other hand they also motivate the need for a collision-free path generation and navigation system for \glspl{uav}. Ensuring a hazard-free, safe \gls{uav} flight is also equally important for indoor applications. Therefore, motion safety of \glspl{uav} has become a prime concern for \gls{uav} operators. It refers to the ability of the \glspl{uav} to detect and avoid collisions with static and moving obstacles in the environment. The static obstacles include buildings, trees, and other similar stationary items, while movable items (for example birds) are considered as moving obstacles.

Some of the commercially-available quadrotors are capable of detecting and avoiding some obstacles. For example, DJI's Phantom~4~Pro\footnote{\url{https://www.dji.com/phantom-4-pro}} uses five-directional sensors to provide obstacle detection or sensing in five directions with a front and rear sensing range of up to 30 meters and up to 7 meters for left and right side. However, its obstacle avoidance mechanism does not work in all kinds of scenarios. In this work, we assume that each \gls{uav} is equipped with an adequate obstacle detection capability and can successfully detect all static and dynamically appearing, moving obstacles in its surroundings. Therefore, the emphasis of this work is not on obstacle detection. Instead, we focus on collision prediction and avoidance.

Multiple \glspl{uav} working in a cooperative manner can be used to provide powerful capabilities that a single \gls{uav} can not offer~\cite{Dong:2009}. Therefore, for larger and highly complex applications and tasks which are either beyond the capabilities of a single \gls{uav} or can not be performed efficiently if only a single \gls{uav} is used, multiple \glspl{uav} can be used together in the form of a swarm or a fleet. In such scenarios, a safe operation can not be guaranteed without preventing the \glspl{uav} from colliding with one another and with static and dynamically appearing, moving obstacles in the flying zone. Therefore, in the context of \gls{uav} swarms, ensuring motion safety entails devising and implementing an online motion path planning, coordination, and navigation system for multiple \glspl{uav} with an integrated support for collision prediction and avoidance.

The problem of motion safety of \glspl{uav} is currently attracting significant research attention. Some comprehensive literature reviews on motion planning algorithms for \glspl{uav} can be found in~\cite{Goerzen:2009, Kendoul:2012}. The main focus of these approaches is on an off-line motion planning phase to plan and produce \gls{uav} paths or trajectories before the start of the mission. Augugliaro et al.~\cite{Augugliaro:2012} also presented a planned approach that generates feasible paths ahead of time. LaValle~\cite{Lavalle:98} and Karaman and Frazzoli~\cite{Karaman:2011} presented sampling-based path planning algorithms. Silva Arantes et al.~\cite{Silva:Arantes:2017} proposed a path planning approach for critical situations requiring an emergency landing of a \gls{uav}. Dong et al.~\cite{Dong:2009} presented a software platform for cooperative control of multiple \glspl{uav}. B{\"u}rkle et al.\cite{Burkle:2011} proposed a multi-agent system architecture for team collaboration in a swarm of drones. Ivanovas et al.~\cite{Ivanovas:2018} proposed an obstacle detection approach for \glspl{uav}. Olivieri~\cite{Olivieri:2015} and Olivieri and Endler~\cite{Olivieri:2015a} presented an approach for movement coordination of swarms of drones using smart phones and mobile communication networks. Their work focuses on the internal communication of the swarm and does not provide a solution for collision-free path generation.

In this paper, we present an online, collision-free path generation and navigation system for swarms of \glspl{uav}. The proposed system uses geographical locations of the \glspl{uav} and of the successfully detected, static and dynamically appearing, moving obstacles to predict and avoid: (1) \gls{uav}-to-\gls{uav} collisions, (2) \gls{uav}-to-static-obstacle collisions, and (3) \gls{uav}-to-moving-obstacle collisions. It comprises three main components: (1) a \gls{cep} and collision prediction module, (2) a mutually-exclusive locking mechanism, and (3) a collision avoidance mechanism. The \gls{cep} and collision prediction module leverages efficient runtime monitoring and \gls{cep} to make timely predictions. The mutually-exclusive locking mechanism prevents multiple \glspl{uav} from attempting to fly to the same location at the same time. The collision avoidance mechanism tries to find best ways to prevent the \glspl{uav} from colliding into one another and with the successfully detected static and moving obstacles in the flying zone. A distinctive feature of the proposed system is its ability to foresee potential collisions and proactively find best ways to avoid the predicted collisions in order to ensure safety of the entire swarm. In contrast to the existing works~\cite{Goerzen:2009, Kendoul:2012, Augugliaro:2012, Olivieri:2015, Olivieri:2015a, Silva:Arantes:2017, Majd:2018:CEC, Majd:2018:PDP, Dong:2009, Burkle:2011, Ivanovas:2018, Karaman:2011, Lavalle:98}, our proposed system does not depend on a planning phase and produces efficient, collision-free paths in an online manner. We focus on collision prediction and avoidance and online path generation and navigation for swarms of \glspl{uav}.

We also present a simulation-based implementation of the proposed system along with an experimental evaluation involving a series of experiments and compare our results with the results of four existing approaches~\cite{Karaman:2011, Lavalle:98, Sujit:2009, Silva:Arantes:2017}. The results show that the proposed system successfully predicts and avoids all three kinds of collisions in an online manner. Moreover, it generates safe and efficient \gls{uav} routes, efficiently scales to large-sized problem instances, and is suitable for cluttered flying zones and for scenarios involving high risks of \gls{uav} collisions. Our proposed navigation system, its implementation, experiments, and results are not based on or limited to a particular application of \gls{uav} swarms. Instead, they are generic enough to be applicable to a wide range of applications. The work presented in this paper extends our preliminary approach and results reported in~\cite{Ashraf:2017:ECBS}.

We proceed as follows. Section~\ref{sec:preliminaries} sets up the terminology and context. The proposed online, collision-free path generation and navigation system for \gls{uav} swarms is presented in Section~\ref{sec:system}. In Section~\ref{sec:example}, we illustrate the main steps of our proposed approach on a small example. Section~\ref{sec:evaluation} presents some important implementation details along with the experimental evaluation. Section~\ref{sec:related_work} reviews important related works. Finally, we present our conclusions in Section~\ref{sec:conclusion}.

\section{Preliminaries}
\label{sec:preliminaries}

The proposed system not only provides support for online collision prediction and avoidance, it also generates complete routes for all \glspl{uav} in the swarm. Unlike traditional motion path planning approaches that require that all obstacles and their precise locations must be known before the start of the mission, the proposed approach does not assume any a priori knowledge of the obstacles. In other words, we assume that the terrain of the flying zone is not known beforehand. Therefore, the proposed system does not make any assumptions on the number and locations of the static and dynamically appearing, moving obstacles. It does not require a preliminary, off-line motion planning phase to produce efficient routes for the \glspl{uav}. In our approach, the drones takeoff from their start locations and fly uninterruptedly towards their destinations until the proposed system predicts a collision and triggers our collision avoidance mechanism to prevent the predicted collision. Since the proposed system uses geographical locations of the \glspl{uav} to generate their paths and predict and avoid collisions, it requires correct and precise location information of all \glspl{uav} in the fleet. Imprecise and incorrect information can result in longer paths and in the worst case some \glspl{uav} can collide with other \glspl{uav} or with some static or moving obstacles.

Let the mission flying zone be represented by a finite set of locations $AREA = \{l_1, l_2, l_3, ..., l_M\}$, where each location $l_i$ is represented as a point in a three-dimensional space $(x, y, z)$. In an outdoor mission, the dimensions $x, y, z$ may correspond to latitude, longitude, and altitude or elevation. To ensure a suitable formation of the swarm, it is assumed that the distance between any two consecutive locations in $AREA$ is less than or equal to the sensing range $sen_r$ of the \glspl{uav} and greater than or equal to the safe distance $dis_s$ for the \glspl{uav}. For example, the front and rear sensing range $sen_r$ of Phantom~4~Pro \gls{uav} is up to 30 meters. Therefore, if the swarm comprises Phantom~4~Pro \glspl{uav}, the maximum distance between any two consecutive locations $l_i, l_j \in AREA \; | \; i \neq j$ should be less than or equal to 30 meters. The safe distance $dis_s$ for \glspl{uav} depends on their maximum speed $Sp$, obstacle detection and processing time $Pt$, and wireless communication latency $Cl$~\cite{Olivieri:2015}. For example, if two \glspl{uav} are found heading towards each other at a maximum speed $Sp$ of 5 meter per second each and with an obstacle detection and processing time $Pt$ of 0.5 seconds and a wireless communication latency $Cl$ of 0.2 seconds, the safe distance $dis_s$ can be estimated as
\begin{equation}
\label{safe_distance}
  dis_s = 2 \cdot Sp \; (2 \cdot Cl + Pt)
\end{equation}
which yields 9 meters. Therefore, in this example, the minimum distance between any two consecutive locations $l_i, l_j \in AREA \; | \; i \neq j$ should be greater than or equal to 9 meters. As a simplification to the problem, we assume that all consecutive locations in $AREA$ are a uniform, fixed distance apart from one another denoted as $dis$, such that $dis_s \leq dis \leq sen_r$. Hence, the flying zone $AREA$ can be viewed as a three-dimensional grid. This simplification allows faster generation, comparison, and evaluation of solutions or \gls{uav} paths. For clarity, important terminology and notation used in this paper are summarized in Table~\ref{tab:notation}.

\begin{table}[!b]
\caption{Terminology and notation} 
\label{tab:notation}
\centering
\begin{tabular}{|l|p{8.0cm}|}
\hline
\textbf{Notation} & \textbf{Description} \\
\hline
$AREA$ & Three-dimensional flying zone \\ \hline
$Cl$ & Wireless communication latency \\ \hline
$dis$ & Distance between two consecutive \glspl{uav} \\ \hline
$dis_s$ & Safe distance for the \glspl{uav} \\ \hline

$SWARM$ & Swarm of drones \\ \hline

$l_{fin}$ & Final or destination location of a \gls{uav} \\ \hline
$l_i$ & A location in $AREA$ \\ \hline
$l_{in}$ & Initial or start location of a \gls{uav} \\ \hline

$MOV\_OBS$ & Set of moving obstacles \\ \hline
$Pt$ & Obstacle detection and processing time \\ \hline

$route_i$ & A \gls{uav} route \\ \hline

$sen_r$ & Sensing range of the \glspl{uav} \\ \hline
$Sp$ & Maximum speed of the \glspl{uav} \\ \hline
$STA\_OBS$ & Set of static obstacles \\ \hline

\end{tabular}
\end{table}

Furthermore, let $SWARM = \{d_1, d_2, ..., d_N\}$ be a set of drones or \glspl{uav} in the swarm. The static obstacles are denoted as $STA\_OBS = \{so_1, so_2, ..., so_O\}$. Similarly the dynamically appearing, moving obstacles are represented by the set $MOV\_OBS = \{mo_1, mo_2, ..., mo_P\}$. Each drone occupies a certain location in $AREA$. The drones takeoff from their start locations and fly towards their destination locations. A drone route or path is a sequence of locations from drone's start location to drone's destination location. For a drone $d_i$, $route_i = <l_{in}, ..., l_{fin}>$ such that $ran(route_i) \subseteq AREA$ and where $l_{in}$ is the initial or start location and $l_{fin}$ is the final or destination location of $d_i$. Similarly, each static and moving obstacle occupies a certain location in $AREA$. Moreover, the moving obstacles keep on moving arbitrarily until they leave the flying zone.

We formulate three basic \glspl{sr} for a swarm of drones:
\begin{description}
  \item[\textbf{\gls{sr}1:}] $\forall d_i \in SWARM, \forall so_j \in STA\_OBS$, $d_i$ does not collide with $so_j$. 
  \item[\textbf{\gls{sr}2:}] $\forall d_i, d_j \in SWARM \; | \; i \neq j$, $d_i$ and $d_j$ do not collide with each other. 
  \item[\textbf{\gls{sr}3:}] $\forall d_i \in SWARM, \forall mo_j \in MOV\_OBS$, $d_i$ does not collide with $mo_j$. 
\end{description}
Since the proposed system does not assume any a priori knowledge on the numbers and locations of the static and moving obstacles and does not depend on a preliminary, off-line motion planning phase, none of the \glspl{sr} can be verified before the start of the mission. For \textbf{\gls{sr}1} which concerns static obstacles, it is necessary that the drones do not fly into a location where a static obstacle is situated. Our proposed system helps the drones to avoid all successfully detected static obstacles in an online manner by providing efficient collision prediction and collision avoidance mechanisms. Similarly, for \textbf{\gls{sr}2} which concerns collisions with other drones, it is required that at any given time $t$ each location is occupied by at most one drone. The proposed system stops the drones from flying into other drones in the vicinity. The proposed mutually-exclusive locking and collision avoidance mechanisms prevent the drones from flying into any locations occupied by other drones at time $t$. For \textbf{\gls{sr}3} which concerns collisions with dynamically appearing, moving obstacles, the proposed system provides a similar approach as for \textbf{\gls{sr}1} that helps the drones to avoid all successfully detected moving obstacles in an online manner.

\section{Collision-Free Path Generation and Navigation}
\label{sec:system}

\begin{figure}[!ht]
\centering
\includegraphics[trim=0 185 180 0, clip, width=0.98\textwidth]{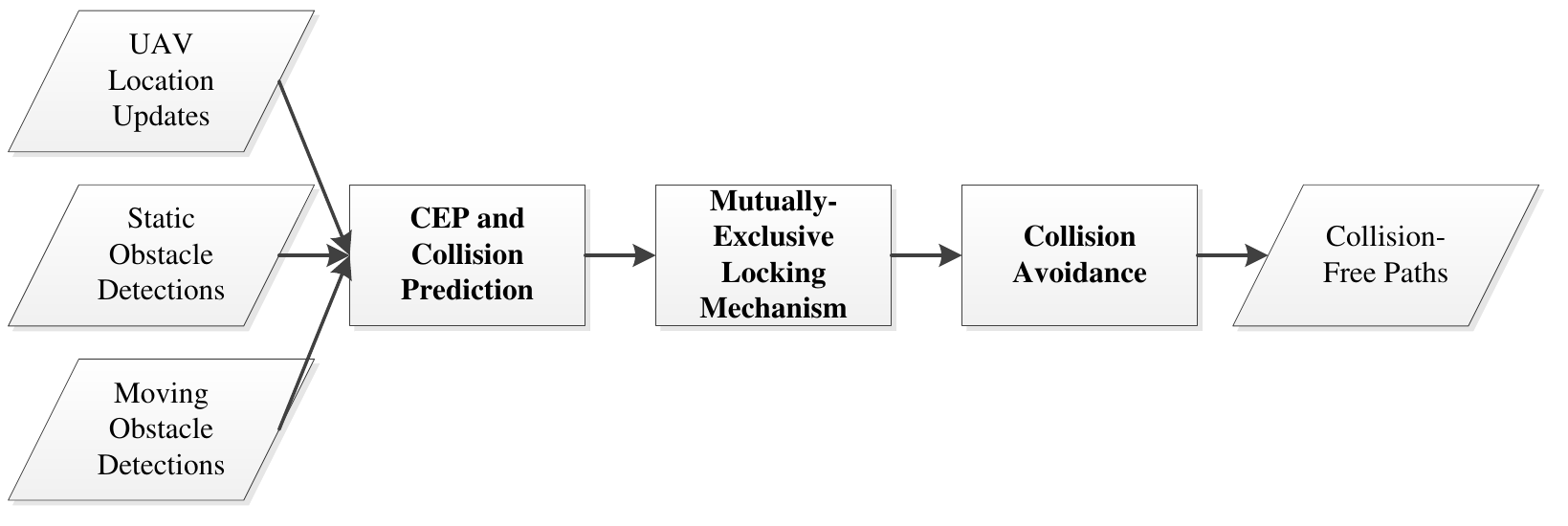}
\caption{Overview of the proposed online, collision-free path generation and navigation system for swarms of \glsentrytext{uav}s}
\label{fig:overview}
\end{figure}

Figure~\ref{fig:overview} presents a high-level system architecture and overview of the proposed online, collision-free path generation and navigation system for swarms of \glspl{uav}. The main components of the proposed system include: (1) a \gls{cep} and collision prediction module, (2) a mutually-exclusive locking mechanism, and (3) a collision avoidance mechanism. The inputs to the system are the \gls{uav} location updates, static obstacle detections, and moving obstacle detections. Based on these three inputs, the \gls{cep} and collision prediction module predicts: (1) \gls{uav}-to-\gls{uav} collisions, (2) \gls{uav}-to-static-obstacle collisions, and (3) \gls{uav}-to-moving-obstacle collisions. Our collision avoidance mechanism tries to find best ways to avoid or bypass collisions and computes collision-free routes for \glspl{uav} in an online manner. In a densely populated and cluttered flying zone, it might not be possible to immediately compute a bypass route for all drones. In such scenarios, the proposed system might put some of the drones into the hover-in-place mode until the situation improves and the routes clear. Additionally, it may also let some \glspl{uav} to temporarily retreat or backtrack to find more suitable, collision-free routes.

The proposed system implements a safety-first approach. Therefore, a hazard-free, safe operation of the swarm takes precedence over all of the other objectives including lengths of the \gls{uav} routes, timely arrival of the \glspl{uav} to their destinations, and achievement of any other mission-specific goals. As a consequence, we do not formulate the problem as an optimization problem. Instead, we implement a stochastic, greedy approach that tries to find safe and efficient routes. The \glspl{uav} takeoff from their start locations and fly uninterruptedly towards their destinations until the \gls{cep} and collision prediction module predicts a collision, in which case our collision avoidance mechanism is invoked to avoid the collision. In addition, our mutually-exclusive locking mechanism prevents multiple \glspl{uav} from attempting to fly to the same location at the same time. At each step, the proposed approach makes a stochastic, greedy decision for each \gls{uav}. It tries to find a next location for each \gls{uav} which is not only safe but also reduces the distance from the destination. The main components of the proposed system are described in the following subsections.

\subsection{Complex Event Processing and Collision Prediction}
\label{sec:cep}

Complex Event Processing (CEP) is a technique for realtime, fast processing of a large number of events from one or more event streams to derive and identify important complex events and patterns in the event streams. \Gls{cep} has been successfully used in a variety of business domains including retail management, health-care, and cloud computing~\cite{Wu:2006, Mdhaffar:2017}. For example in retail management, \gls{cep} can be used to detect shoplifting and out-of-stock events. The basic or primitive events in \gls{cep} are processed into complex or composite events by means of event processing queries, which are written in a \gls{sql}-like language. Therefore, \gls{cep} provides a similar functionality for realtime event streams that a relational database management system provides for persistent data.

One of the most widely used \gls{cep} tools is the Esper \gls{cep} engine\footnote{\url{http://www.espertech.com/esper/}}, in which the event processing queries are written in the \gls{epl}. There are three main steps for using Esper \gls{cep} engine:
\begin{enumerate}
  \item In the first step, event types and sources of events are registered with the \gls{cep} engine. An event class in Esper is written as a \gls{pojo}.
  \item The second step requires event processing queries to be written in \gls{epl}.
  \item Finally, in the third step, event sinks are implemented which can be used to perform some suitable control and repair actions.
\end{enumerate}

The \gls{cep} and collision prediction module in our proposed system uses a \gls{cep} engine to monitor and keep track of the current location of the \glspl{uav} and of the successfully detected static and moving obstacles. Table~\ref{tab:cep_events} presents the three types of events from the proposed system along with their properties. The \glspl{uav} generate and send location update events on regular intervals, for example every 50 milliseconds. A drone location event (\emph{DroneLocEvent}) contains drone name of the concerned drone $d_i \in SWARM$, drone location $l_i$ in the three-dimensional flying zone $AREA$, and the event time $t$. The \gls{cep} engine receives and processes these events to predict possible \gls{uav}-to-\gls{uav} collisions in the swarm. Similarly, for each successfully detected static obstacle, a static obstacle event (\emph{SObsEvent}) is generated and sent to the \gls{cep} engine. A static obstacle detection event contains obstacle name of the static obstacle $so_i \in STA\_OBS$ and the location $l_i \in AREA$ of the static obstacle. The \gls{cep} engine processes all \gls{uav} location update events and static obstacle detection events to predict \gls{uav}-to-static-obstacle collisions. Finally, for successfully detected moving obstacles, moving obstacle events (\emph{MObsEvents}) are generated and sent to the \gls{cep} engine. A moving obstacle detection event contains obstacle name of the moving obstacle $mo_i \in MOV\_OBS$, the location $l_i \in AREA$ of the moving obstacle, and the event time $t$. The \gls{cep} engine processes \gls{uav} location update events and moving obstacle detection events to predict \gls{uav}-to-moving-obstacle collisions.

\begin{table}[!t]
\caption{Three types of CEP events from the proposed system}
\label{tab:cep_events}
\centering
\begin{tabular}{|l|l|} 
\hline
\textbf{Event type} & \textbf{Properties} \\
\hline
\emph{DroneLocEvent} & Drone name $d_i$, drone location $l_i$, event time $t$ \\ \hline
\emph{SObsEvent}  & Static obstacle name $so_i$, obstacle location $l_i$ \\  \hline
\emph{MObsEvent} & Moving obstacle name $mo_i$, obstacle location $l_i$, event time $t$ \\ \hline
\end{tabular}
\end{table}

The proposed system implements three \gls{epl} queries to process the three types of events and determine if a drone is flying in a close proximity of another drone or a static or moving obstacle. Listing~\ref{list:epl1} presents the first query. It uses DroneLocEvents to check if two drones are in a close proximity of each other. If a match is found, the \gls{cep} engine triggers the concerned event sink, which may predict a \gls{uav}-to-\gls{uav} collision and then invoke the collision avoidance mechanism to prevent the \glspl{uav} from colliding into each other.

\lstset{language=Java, frame=single, caption={EPL query to determine if two drones are in a close proximity of each other}, label=list:epl1, basicstyle=\normalsize\ttfamily}
\begin{center}
\begin{minipage}{0.95\linewidth}
\centering
\begin{lstlisting}
select A.droneName as aName, A.x as aX, A.y as aY, A.z as aZ,
B.droneName as bName, B.x as bX, B.y as bY, B.z as bZ, from
DroneLocEvent.win:time(1 sec) A, DroneLocEvent.win:time(1 sec)
B where A.droneName != B.droneName and A.x in [B.x-2:B.x+2]
and A.y in [B.y-2:B.y+2] and A.z in [B.z-2:B.z+2]
and (A.x = B.x or A.y = B.y or A.z = B.z)
\end{lstlisting}
\end{minipage}
\end{center}

The second query in Listing~\ref{list:epl2} uses DroneLocEvents and SObsEvents to determine if a drone is in close proximity of a static obstacle. Similarly, the third query in Listing~\ref{list:epl3} uses DroneLocEvents and MObsEvents to determine if a drone is in close proximity of a moving obstacle. In each case, the relevant event sink is triggered, which may predict a collision and invoke the collision avoidance mechanism.

\lstset{language=Java, frame=single, caption={EPL query to determine if a drone is in close proximity of a static obstacle}, label=list:epl2, basicstyle=\normalsize\ttfamily}
\begin{center}
\begin{minipage}{0.95\linewidth}
\centering
\begin{lstlisting}
select A.droneName as aName, A.x as aX, A.y as aY, A.z as aZ,
O.obstacleName as oName, O.x as oX, O.y as oY, O.z as oZ, from
DroneLocEvent.win:time(1 sec) A, SObsEvent.win:time(1 hour) O
where A.x in [O.x-1:O.x+1] and A.y in [O.y-1:O.y+1] and A.z in
[O.z-1:O.z+1] and (A.x = O.x or A.y = O.y or A.z = O.z)
\end{lstlisting}
\end{minipage}
\end{center}

\lstset{language=Java, frame=single, caption={EPL query to determine if a drone is in close proximity of a moving obstacle}, label=list:epl3, basicstyle=\normalsize\ttfamily}
\begin{center}
\begin{minipage}{0.95\linewidth}
\centering
\begin{lstlisting}
select A.droneName as aName, A.x as aX, A.y as aY, A.z as aZ,
O.obstacleName as oName, O.x as oX, O.y as oY, O.z as oZ, from
DroneLocEvent.win:time(1 sec) A, MObsEvent.win:time(1 sec) O
where A.x in [O.x-2:O.x+2] and A.y in [O.y-2:O.y+2] and A.z in
[O.z-2:O.z+2] and (A.x = O.x or A.y = O.y or A.z = O.z)
\end{lstlisting}
\end{minipage}
\end{center}

To predict the three different kinds of collisions, the event sinks use various parameters and rules. The parameters include the (current) locations of the drones and of the static and moving obstacles and the desired next locations of the drones. The collision prediction rules are presented in Algorithm~\ref{algo:rules}. Rule 1 states that a \gls{uav}-to-\gls{uav} collision is predicted when the desired next location of a drone $d_i \in SWARM$ is same as the current or the desired next location of another drone $d_j \in SWARM \; | \; i \neq j$. Similarly, Rule 2 is used for predicting \gls{uav}-to-static-obstacle collisions, which can occur if a drone $d_i \in SWARM$ attempts to fly to a location occupied by a static obstacle $so_j \in STA\_OBS$. Finally, Rule 3 states that a \gls{uav}-to-moving-obstacle collision is predicted when the desired next location of a drone $d_i \in SWARM$ is same as the current location of a moving obstacle $mo_j \in MOV\_OBS$.

\begin{algorithm}[!b]
\caption{Collision prediction rules}
\label{algo:rules}
\centering
\begin{algorithmic}[1]

\STATE\COMMENT{\textbf{Rule 1}}
\STATE $\forall d_i, d_j \in SWARM \; | \; i \neq j$, let $l_i, l_m \in AREA$ be the current and the desired next locations of $d_i$ and similarly $l_j, l_n \in AREA$ be the current and the desired next locations of $d_j$
\IF {$l_m = l_j$ $\vee$ $l_m = l_n$ $\vee$ $l_n = l_i$}
\STATE predict a \gls{uav}-to-\gls{uav} collision
\ENDIF

\STATE\COMMENT{\textbf{Rule 2}}
\STATE $\forall d_i \in SWARM, \forall so_j \in STA\_OBS$, let $l_m \in AREA$ be the desired next location of $d_i$ and $l_j \in AREA$ be the location of $so_j$
\IF {$l_m = l_j$}
\STATE predict a \gls{uav}-to-static-obstacle collision
\ENDIF

\STATE\COMMENT{\textbf{Rule 3}}
\STATE $\forall d_i \in SWARM, \forall mo_j \in MOV\_OBS$, let $l_m \in AREA$ be the desired next location of $d_i$ and $l_j \in AREA$ be the current location of $mo_j$
\IF {$l_m = l_j$}
\STATE predict a \gls{uav}-to-moving-obstacle collision
\ENDIF

\end{algorithmic}
\end{algorithm}

\subsection{Mutually-Exclusive Locking Mechanism}
\label{sec:locks}

The \gls{cep} and collision prediction module described in the previous section covers most of the scenarios that can lead to a \gls{uav} collision. However, since the \glspl{uav} may move fast and arbitrarily, some failures and collisions can still occur. For instance, during a mission, a location $l_i \in AREA$ is free and two \glspl{uav} $d_i, d_j \in SWARM \; | \; i \neq j$ concurrently decide to move to $l_i$. If the \gls{cep} and collision prediction module takes slightly longer to predict the \gls{uav}-to-\gls{uav} collision, the collision avoidance mechanism might not be left with enough time to prevent the collision. However, if the \gls{cep} and collision prediction module quickly and correctly predicts the collision, the collision avoidance module can save the \glspl{uav} $d_i, d_j$ by allowing only one of them to continue flying to $l_i$. The other \gls{uav} will either be redirected to another location or will fail to move in the current iteration. To prevent such scenarios and failures, we augment our collision prediction approach with a mutually-exclusive locking mechanism.

The proposed system uses mutually-exclusive locks on the current and the immediate next locations of \glspl{uav} to prevent multiple \glspl{uav} from attempting to move to the same location at the same time. The lock state of each location $l_i \in AREA$ can be either \emph{locked} or \emph{unlocked}. The current location of each \gls{uav} is always considered locked for all other \glspl{uav}. Moreover, as soon as a \gls{uav} decides its next move, the system puts a mutually-exclusive lock on the immediate next location of the \gls{uav} so that the other \glspl{uav} do not attempt to move to the same location. Similarly, while deciding about their next moves, the \glspl{uav} first check the lock state of the possible next locations and only attempt to move to some of the unlocked locations. Moreover, if multiple \glspl{uav} $d_1, d_2, ..., d_N \in SWARM$ concurrently attempt to lock the same location, only one of them acquires the lock. For scenarios involving very short time intervals, the mutually-exclusive locks may be acquired in one time interval and the moves may be performed in the next time interval. The \glspl{uav} release the locks of their previous locations as soon as they fly to their next locations.

\begin{figure}[!h]
\centering
\includegraphics[trim=0 400 170 0, clip, width=0.98\textwidth]{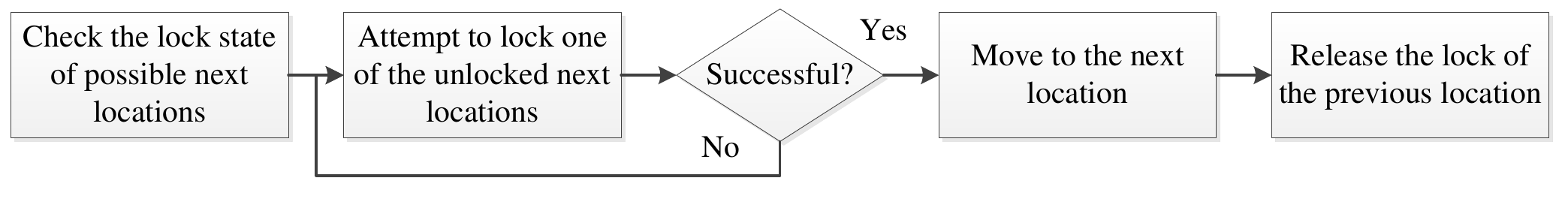}
\caption{The mutually-exclusive locking mechanism}
\label{fig:locks}
\end{figure}

Figure~\ref{fig:locks} illustrates the proposed locking mechanism. It shows that a \gls{uav} always keeps a mutually-exclusive lock for its current location. Moreover, when deciding about a next move, it first checks the lock state of all possible next locations. It then attempts to lock one of the unlocked locations. After successful locking of the next location, it moves to the next location. Finally, it releases the lock of the previous location. In this way, two or more \glspl{uav} never attempt to move to the same location at the same time.

\subsection{Collision Avoidance Mechanism}
\label{sec:avoidance}

Whenever the \gls{cep} and collision prediction module predicts a collision, it invokes our collision avoidance mechanism which tries to find best ways to avoid the predicted collisions and computes collision-free routes for \glspl{uav} in an online manner. Based on the severity of the predicted collision, its surroundings, and the overall situation of the $SWARM$ and of the successfully detected static and moving obstacles ($STA\_OBS$ and $MOV\_OBS$) in $AREA$, our collision avoidance mechanism uses one of the three collision avoidance techniques in the following order: (1) redirecting the \gls{uav} into another direction, (2) putting the \gls{uav} into the hover-in-place mode until the route is cleared, and (3) temporarily retreating or backtracking the \gls{uav} to explore some alternate collision-free routes. The pseudocode of the proposed collision avoidance mechanism is given as Algorithm~\ref{algo:avoidance}. Our backtracking approach is described in the following section.

\begin{algorithm}[!b]
\caption{Collision avoidance mechanism}
\label{algo:avoidance}
\centering
\begin{algorithmic}[1]
\STATE redirect the \gls{uav} into another direction

\IF {not successful}
\STATE activate the hover-in-place mode until the \gls{uav} route is cleared
\ENDIF

\IF {the \gls{uav} hovers for too long or if it takes too long to find a suitable collision-free route}
\STATE temporarily backtrack the \gls{uav} to explore some alternate collision-free routes
\ENDIF
\end{algorithmic}
\end{algorithm}

The first collision avoidance technique namely redirecting the \gls{uav} into another direction means changing the flying direction of the \gls{uav}. For example, if a \gls{uav} is flying in the $x$ dimension of $AREA$, but the \gls{cep} and collision prediction module predicts a collision due to the presence of an obstacle or another \gls{uav} on the path, then the \gls{uav} can not continue a hazard-free flight in the $x$ dimension any more. Therefore, the collision avoidance mechanism redirects the \gls{uav} to fly in the $y$ or $z$ dimension so the \gls{uav} may be able to avoid the collision. However, in a densely populated and cluttered flying zone, the collision avoidance mechanism might not be able to immediately compute a bypass route for all drones. Therefore, in such scenarios, the proposed collision avoidance mechanism activates the hover-in-place mode for some of the \glspl{uav} until the situation improves and the routes clear. Additionally and as a last resort, it temporarily backtracks some \glspl{uav} to explore some alternate collision-free routes. It should be noted that all three collision avoidance techniques incur some overhead, which might extend the routes and increase the flight durations for some of the \glspl{uav}. However, as explained previously, this is inevitable for a safety-first approach.

\subsection{Backtracking Approach}
\label{sec:backtracking}

The proposed backtracking approach temporarily retreats or backtracks a \gls{uav} so it may explore some alternate collision-free routes. As shown in Algorithm~\ref{algo:avoidance}, the proposed backtracking algorithm is triggered in two situations: (1) if a \gls{uav} hovers for too long or (2) if a \gls{uav} keeps on moving but it does not find a suitable collision-free route to reach to its destination. In densely populated, cluttered flying zones, such situations are not unprecedented. Sometimes, a \gls{uav} keeps on hovering or moving near its destination, but it does not find a collision-free route to reach to the destination because some other \glspl{uav} or obstacles reside between the \gls{uav} and its destination location and thus obstructs the \gls{uav}'s routes. In such situations, it is important to allow the \gls{uav} to temporarily retreat or backtrack so it may be able to explore some alternate routes to reach to its destination.

The pseudocode of the proposed backtracking algorithm is presented as Algorithm~\ref{algo:backtrack}. The algorithm iterates until the required number of backtrack steps is successfully completed or the maximum number of backtrack attempts is reached (line~9). In each iteration, it attempts to move the \gls{uav} in the opposite direction of the \gls{uav}'s destination. It randomly chooses one of the three dimensions $(x, y, z)$ and tries to move the \gls{uav} so that the distance from the destination is increased. The backtrack algorithm does not disable the \gls{cep} and collision prediction module and the mutually-exclusive locking mechanism. Thus, in each iteration, the \gls{uav} either backtracks one step or if a collision-hazard is found or the \gls{uav} fails to lock the required location then it hovers at its current location. When the \gls{uav} returns to the normal flight mode (line~10), it explores some alternate collision-free routes to reach to its destination.

\begin{algorithm}[!b]
\caption{Backtracking}
\label{algo:backtrack}
\centering
\begin{algorithmic}[1]
\WHILE {the \gls{uav} is in the backtrack mode}
\STATE randomly choose one of the three dimensions $(x, y, z)$
\STATE attempt to move the \gls{uav} in the opposite direction of its destination
\IF {a collision-hazard is found or the required location is locked by another \gls{uav}}
\STATE hover-in-place
\ELSE
\STATE backtrack the \gls{uav} by moving it one step away from its destination
\ENDIF
\IF {the required number of backtrack steps is successfully completed or the maximum number of backtrack attempts is reached}
\STATE deactivate the backtrack mode and return to the normal flight mode
\ENDIF
\ENDWHILE
\end{algorithmic}
\end{algorithm}

\section{An Illustrative Example}
\label{sec:example}

In this section, we present a small example to illustrate the main components and steps of the proposed online, collision-free path generation and navigation system. Although the proposed system works for a realistic, three-dimensional flying zone, it is difficult to illustrate and demonstrate a three-dimensional flying zone on a paper. Therefore, we use a two-dimensional flying zone for a simpler illustration.

\begin{figure*}[b!]
    \centering
    \subfloat[Before the start of the mission]{\includegraphics[trim=20 110 280 10, clip, width=0.45\textwidth]{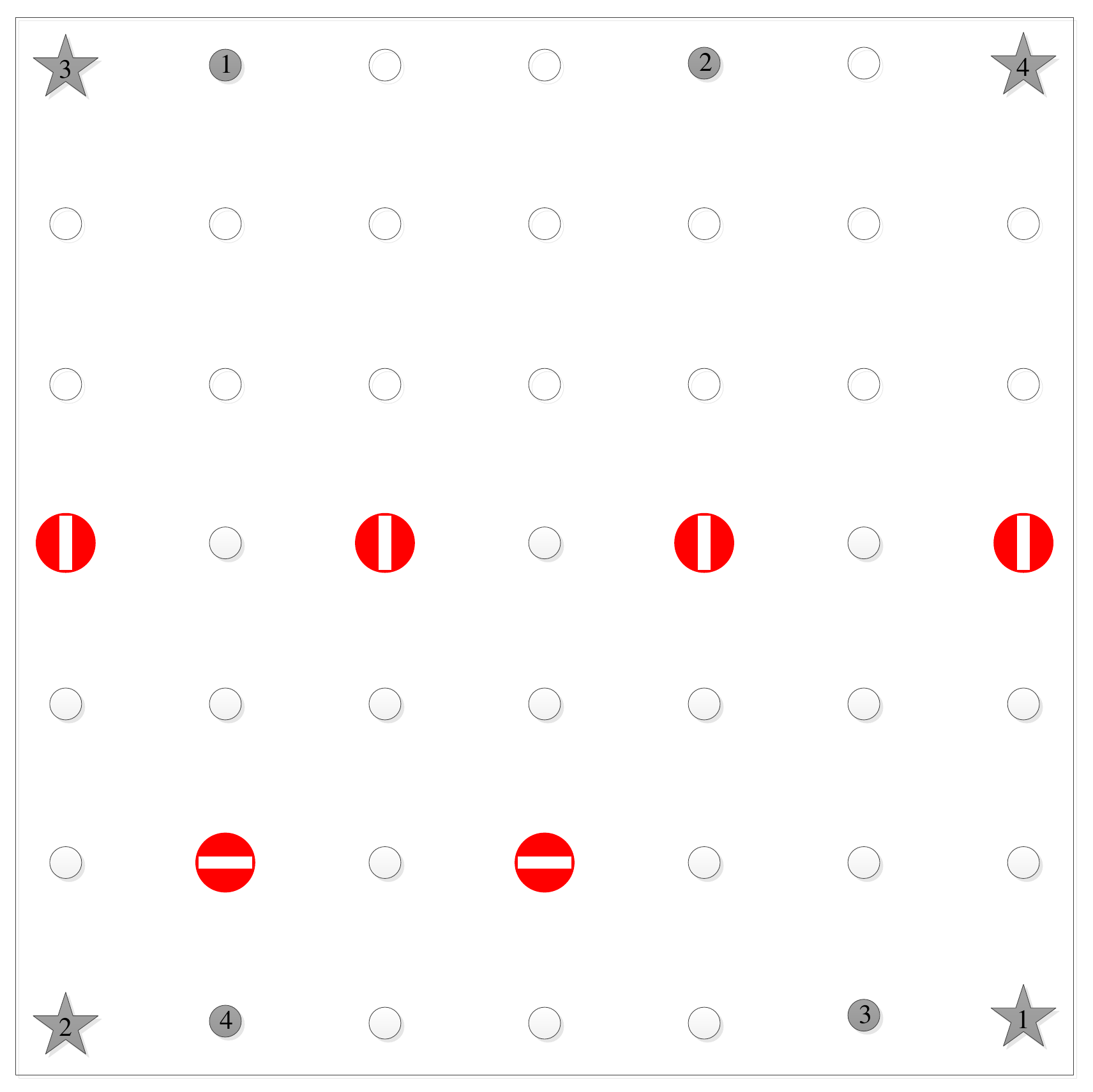}\label{fig:example1}}
    \hspace{0.001\textwidth}
    \subfloat[After five time intervals]{\includegraphics[trim=20 110 280 10, clip, width=0.45\textwidth]{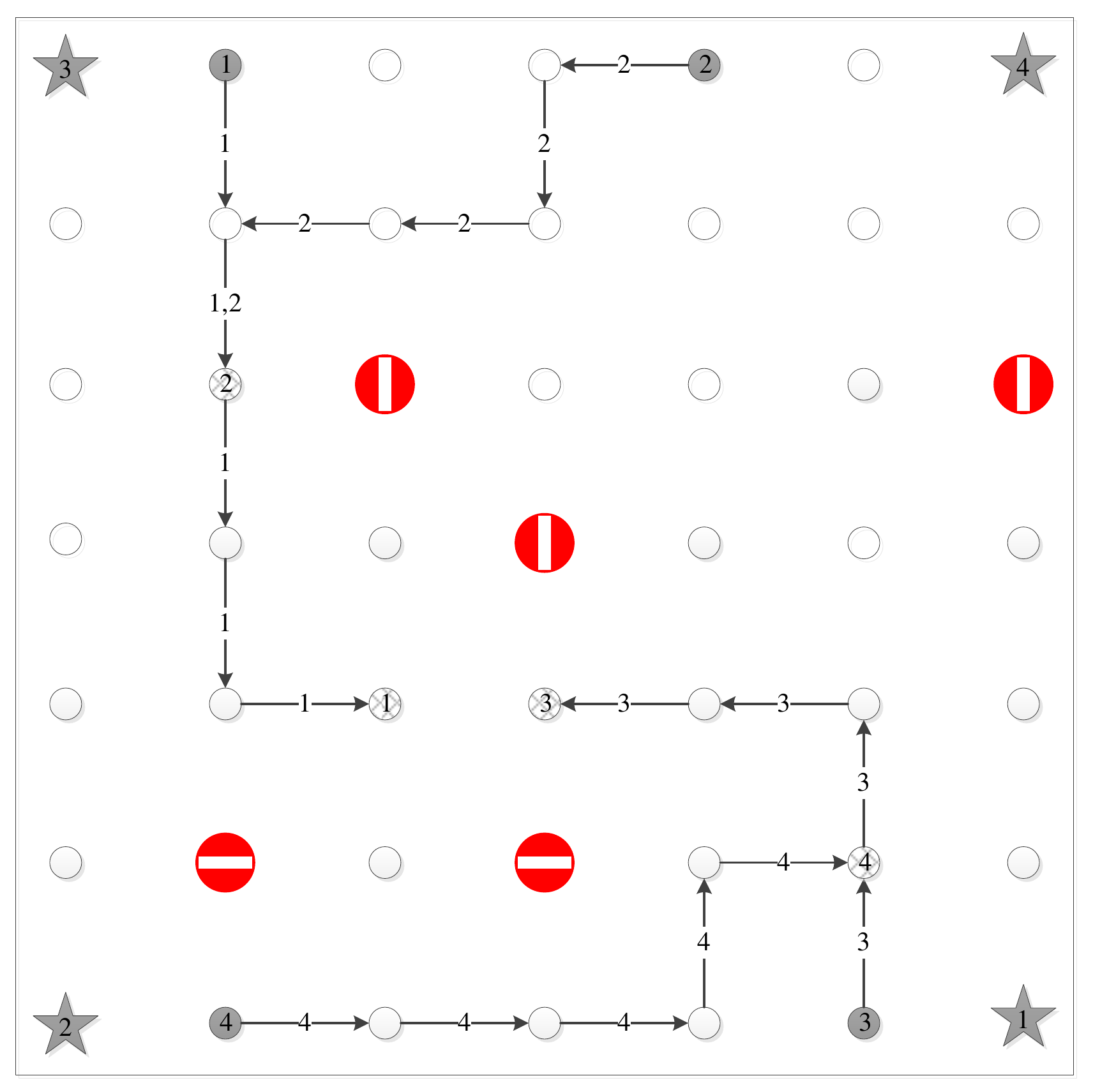}\label{fig:example2}}
    \hspace{0.001\textwidth}
    \subfloat[After the completion of the mission]{\includegraphics[trim=20 110 280 10, clip, width=0.45\textwidth]{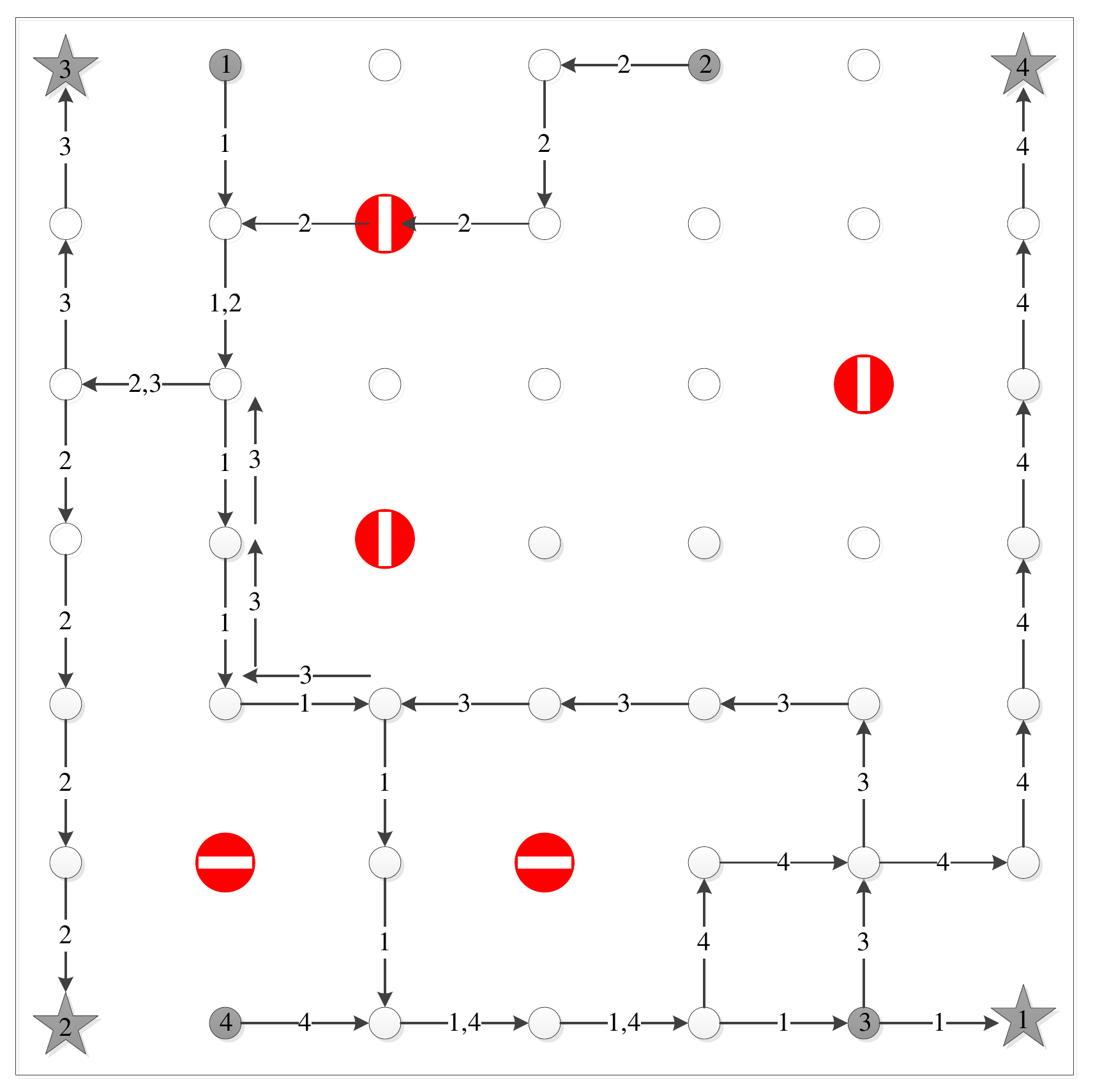}\label{fig:example3}}
    \hspace{0.001\textwidth}
    \subfloat{\includegraphics[trim=0 300 465 0, clip, width=0.45\textwidth]{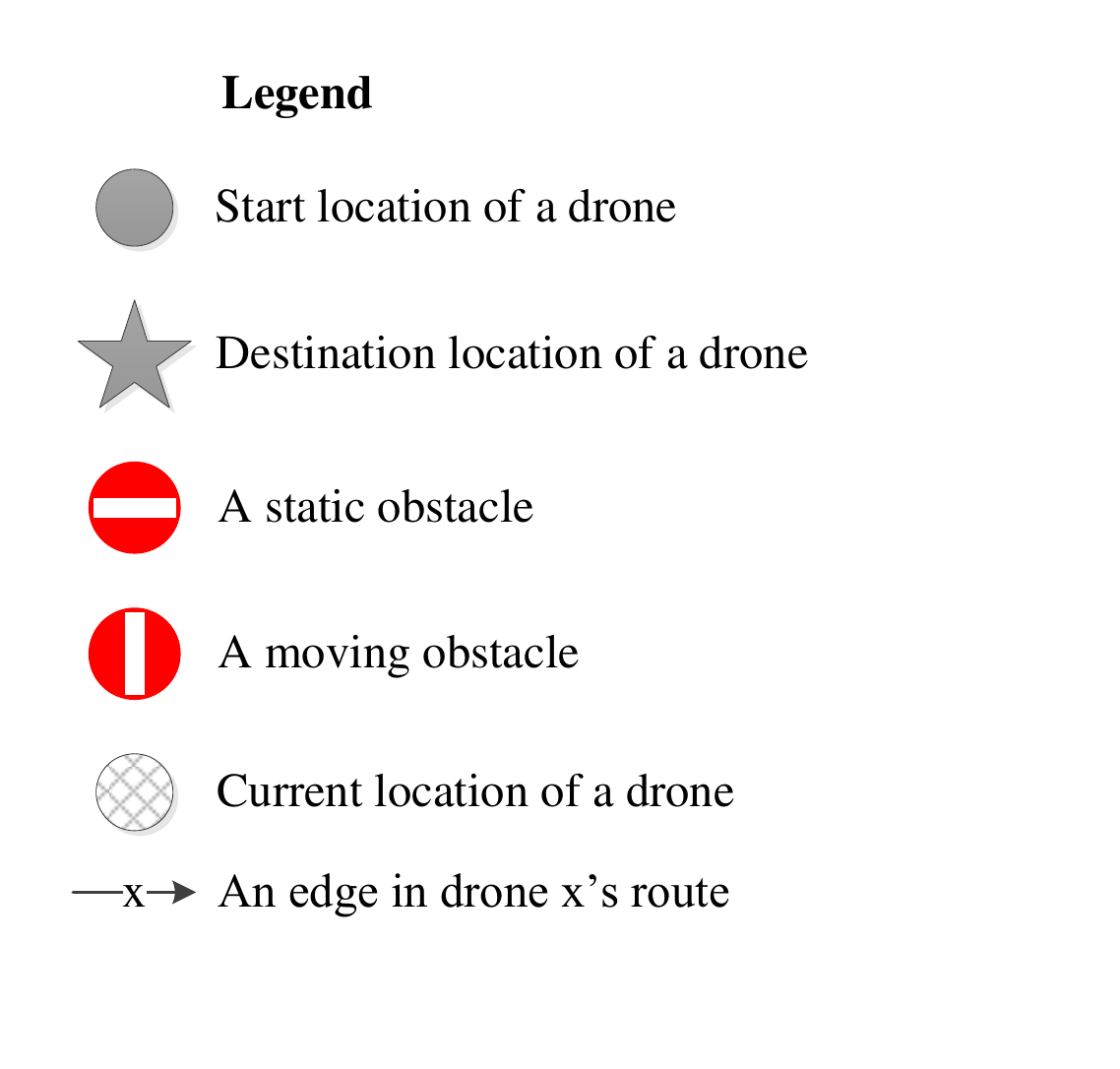}}
    \caption{A simple illustrative example with four \glsentrytext{uav}s, two static obstacles, and four moving obstacles in a two-dimensional flying zone}
    \label{fig:example}
\end{figure*}

Figure~\ref{fig:example} presents an illustrative example with four \glspl{uav}, two static obstacles, and four moving obstacles in a two-dimensional flying zone. The flying zone in our example is shown as a 7x7 grid, where all consecutive locations are a uniform, fixed distance apart from one another. The start and destination locations of each drone are also highlighted. The goal is to route the drones from their start locations to their destination locations while avoiding collisions with static and moving obstacles and with the other drones in the swarm.

It should be noted that the knowledge of the precise locations of the obstacles in this example is only for illustration purposes. As described previously, the proposed system does not make any assumptions on the number and locations of the static and moving obstacles in the flying zone. Similarly, although Figure~\ref{fig:example1} shows that all moving obstacles are present in the flying zone before the start of the mission, in a realistic scenario some moving obstacles (for example birds) may dynamically appear in the flying zone during the execution of the mission.

Figure~\ref{fig:example2} presents a snapshot of the flying zone after five time intervals have elapsed since the start of the mission. It shows that each \gls{uav} started flying from its start location and flew towards its destination location while randomly choosing to fly in the horizontal or vertical dimension in each time interval. Figure~\ref{fig:example2} also shows that the left most moving obstacle from Figure~\ref{fig:example1} left the flying zone during the execution of the mission and that the remaining moving obstacles moved to some new arbitrary locations within the flying zone. Although the moving obstacles moved in an arbitrary fashion either horizontally or vertically, in five time intervals each moving obstacle moved only one step, that is, only to a next consecutive location in the flying zone. Therefore, the moving obstacles moved slower than the drones. This is a reasonable assumption because if the moving obstacles move faster than the drones, even the most advanced and fastest collision detection, prediction, and avoidance mechanisms will not be able to avoid \gls{uav}-to-moving-obstacle collisions.

The labelled, directional edges in Figure~\ref{fig:example2} show the collision-free \gls{uav} routes generated by the proposed system in an online manner. For example, in the top left corner of Figure~\ref{fig:example2}, the first downward edge labelled 1 means that \gls{uav} 1 flew in the downward direction. Similarly, the next edge in the same direction labelled 1,2 shows that \gls{uav} 1 and 2 used the same edge. However, two \glspl{uav} using the same edge does not mean a \gls{uav}-to-\gls{uav} collision. A \gls{uav}-to-\gls{uav} collision on an edge can happen when two \glspl{uav} fly at the same edge at the same time. In this example, \gls{uav} 1 and \gls{uav} 2 flew on the same edge, but in different time intervals. \Gls{uav} 1 left the edge before \gls{uav} 2 arrived there and hence there was no collision-hazard between the two \glspl{uav}. Figure~\ref{fig:example2} also shows the current locations of the \glspl{uav} after five time intervals. It can be seen that all \glspl{uav} except \gls{uav} 3 flew five steps. \Gls{uav} 3 flew four steps and then hovered in the fifth time interval because the system could not find a collision-free move for \gls{uav} 3.

The \glspl{uav} in Figure~\ref{fig:example} used the proposed mutually-exclusive locking mechanism at each step. As described in Section~\ref{sec:locks}, each \gls{uav} first attempted to lock one of the unlocked next locations. After successful locking of their next locations, the \glspl{uav} moved to their next locations and released the locks of their previous locations. Therefore, at any time, two or more \glspl{uav} did not attempt to move to the same location in the flying zone.

\Gls{uav} 1 in Figure~\ref{fig:example2} started flying vertically in the downward direction and continued towards its destination until it detected a static obstacle. At this stage, our \gls{cep} and collision prediction module predicted a \gls{uav}-to-static-obstacle collision and invoked our collision avoidance mechanism, which redirected the \gls{uav} into the horizontal, rightward direction so the drone could continue flying towards its destination. However, in the same time interval, \gls{uav} 3 tried to fly into the same location where \gls{uav} 1 was headed. The two \glspl{uav} detected each other and the \gls{cep} and collision prediction module predicted a \gls{uav}-to-\gls{uav} collision. As a result, our collision avoidance mechanism was invoked, which tried to redirect \gls{uav} 3 in the vertical, upward direction, but the \gls{uav} detected a moving obstacle at that location and the \gls{cep} and collision prediction module predicted a \gls{uav}-to-moving-obstacle collision. Therefore, the collision avoidance mechanism activated the hover-in-place mode for \gls{uav} 3, but let \gls{uav} 1 to lock and then move to the next location. Hence, \gls{uav} 3 flew only four steps in five time intervals. In this example, \gls{uav} 2 and 4 did not encounter a collision-hazard and flew normally towards their destinations. Moreover, none of the \glspl{uav} hovered for too long or took too long to find a suitable, collision-free route. Therefore, our backtracking algorithm was not invoked.

Figure~\ref{fig:example3} shows a snapshot of the flying zone after the completion of the mission. It shows that how each drone found its way to its destination while avoiding obstacles and other drones on its way. Once again, the remaining three moving obstacles moved to some new arbitrary locations within the flying zone. In the sixth time interval, \gls{uav} 1 was redirected in the downward direction to avoid a collision with \gls{uav} 3. Similarly, after flying downwards for two time intervals, \gls{uav} 1 reached the end of the flying zone and was once again redirected to the horizontal, rightward direction. Finally, after flying for a few more intervals in the rightward direction, \gls{uav} 1 reached its destination. As can be seen in Figure~\ref{fig:example3}, all other \glspl{uav} found their ways in similar ways.

\section{Implementation and Experimental Evaluation}
\label{sec:evaluation}

To demonstrate and evaluate our proposed system, we have developed a software simulator. This section briefly describes some important implementation details along with an experimental evaluation involving a series of experiments. We also compare our results with the results of the following four approaches:
\begin{enumerate}
  \item \textbf{\Gls{pso} based approach:} Sujit and Beard~\cite{Sujit:2009}'s \gls{pso} based path planning approach generates paths for a swarm of drones.
  \item \textbf{Greedy heuristics and \glspl{ga} approach:} Silva Arantes et al.~\cite{Silva:Arantes:2017}'s approach uses greedy heuristics and \glspl{ga} to generate and optimize paths for a \gls{uav} under critical situations.
  \item \textbf{\Gls{rrt}:} A sampling-based path planning algorithm proposed by LaValle~\cite{Lavalle:98}.
  \item \textbf{\Gls{rrt}*:} An extension of \gls{rrt} developed by Karaman and Frazzoli~\cite{Karaman:2011} to plan optimal paths.
\end{enumerate}

\subsection{Implementation Details}

The implementation of the first main component of the proposed system called the \gls{cep} and collision prediction module is based on the Esper \gls{cep} engine and Algorithm~\ref{algo:rules}. The second component, called the mutually-exclusive locking mechanism, implements the proposed locking mechanism presented in Figure~\ref{fig:locks}. Similarly, the collision avoidance component implements Algorithm~\ref{algo:avoidance} and~\ref{algo:backtrack}.

We have implemented a simple, controlled simulation platform that does not take into account complex physical phenomena and uncontrolled environment variables such as gravity and wind. The objective is to test and evaluate the proposed system in an ideal scenario while ignoring and minimizing the effects of external uncontrolled factors. Therefore, it is easier to analyze and interpret the results. The implementation assumes that all drones fly at the same speed and that there were no internal drone failures during the execution of the mission. We also assume that at least one feasible path exists for each \gls{uav}.

\subsection{Experiment Design and Setup}

Table~\ref{tab:exp} presents the experiment design. The experimental evaluation comprises four experiments. In each experiment, we ran our proposed approach, Sujit and Beard~\cite{Sujit:2009}'s \gls{pso} based approach, Silva Arantes et al.~\cite{Silva:Arantes:2017}'s greedy heuristics and \gls{ga} based approach, LaValle~\cite{Lavalle:98}'s \gls{rrt} algorithm, and Karaman and Frazzoli~\cite{Karaman:2011}'s \gls{rrt}* algorithm 10 times and used a random seed every time. All results reported in this section are averaged over 10 runs. The experiments were run on an Intel Core i7-4790 processor with 16 gigabytes of memory. The length of the time interval used in the software simulator was 50 milliseconds. We measured the following dependent variables:
\begin{itemize}
  \item \textbf{\Gls{arl}:} the average \gls{uav} route length measured as the number of \gls{uav} moves in the discretized flying zone. A \gls{uav} route is a sequence of moves or steps from \gls{uav}'s start location to \gls{uav}'s destination location. To be minimized to generate shorter routes.
  \item \textbf{\Gls{llr}:} the total number of steps in the longest generated route. To be minimized to generate shorter routes.
  \item \textbf{\Gls{nc}:} the number of \gls{uav} collisions. To be minimized to generate safer routes.
  \item \textbf{\Gls{t}:} computation time of the algorithm in milliseconds (ms). It is the time that the algorithm takes to run and produce the results. To be minimized to reduce the computation overhead.
\end{itemize}

\begin{table}[!t]
\caption{Experiment design}
\label{tab:exp}
\centering
\begin{tabular}{|l|c|c|c|>{\centering\arraybackslash}p{3.3cm}|}
\hline
\textbf{Experiment} & \textbf{1} & \textbf{2} & \textbf{3} & \textbf{4} \\ 
\hline
Experiment type & Small & Large & Cluttered flying zone & High risk of \gls{uav} collisions \\ \hline
Flying zone & 10x10x10 & 20x20x20 & 10x10x10 & 20x20x20 \\ \hline 
Number of \glspl{uav} & 20 & 50 & 20 & 100 \\ \hline
Static obstacles & 20 & 50 & 40 & 50 \\ \hline
Moving obstacles & 20 & 50 & 40 & 50 \\ \hline
\end{tabular}
\end{table}

Experiment 1 was designed to simulate a small problem instance. The main objective was to evaluate the collision prediction and avoidance capabilities of the proposed system in a simpler scenario. The experiment used a 10x10x10 flying zone with 20 drones, 20 static obstacles, and 20 moving obstacles. Experiment 2 used a large problem instance involving a larger flying zone and a higher number of drones and obstacles and was designed to evaluate the proposed system for a larger problem instance. The experiment used a 20x20x20 flying zone with 50 drones, 50 static obstacles, and 50 moving obstacles.

The third experiment evaluated the proposed system for a densely populated, cluttered environment involving a large number of static and moving obstacles. Experiment 3 used a similar experiment design as Experiment 1, but with twice as many static and moving obstacles. Finally, the objective of Experiment 4 was to evaluate the proposed system in a scenario involving high risks of drone collisions. The experiment used a similar experiment design as Experiment 2, but with twice as many drones.

All drones and obstacles were placed randomly. However, to ensure that the drones do not collide during takeoff, unique start locations were used and no obstacles were placed at the drone start locations. Similarly, the destination locations for the drones were also chosen randomly, but it was ensured that all destination locations are unique and that no obstacles were present at the destination locations.

\subsection{Results and Analysis}

The results are presented in Table~\ref{tab:results}. The best results in the table are highlighted in \textbf{bold} font. The results in the \gls{arl} and \gls{llr} columns show that the \gls{rrt}* algorithm~\cite{Karaman:2011} produced the shortest drone routes in all experiments, while \gls{rrt}~\cite{Lavalle:98} generated the second shortest routes. The \gls{nc} column shows that the proposed approach produced the safest routes in all experiments, while the greedy heuristics and \gls{ga} based approach~\cite{Silva:Arantes:2017} produced the second safest routes. In terms of \gls{t}, \gls{rrt} performed the best, while the proposed approach performed second best in Experiment 2 and 3, and the greedy and \gls{ga} based approach performed second best in Experiment 1 and 4. The \gls{pso} based approach~\cite{Sujit:2009} did not perform best or second best with respect to any dependent variable.

In the first experiment, the \gls{arl} for the proposed, \gls{pso} based, greedy and \gls{ga} based, \gls{rrt}, and \gls{rrt}* algorithms was 17, 25, 26, 16, and 15, respectively. Similarly, the \gls{llr} for the proposed, \gls{pso} based, greedy and \gls{ga} based, \gls{rrt}, and \gls{rrt}* algorithms was 36, 49, 47, 36, and 29, respectively. The five approaches also produced similar results in Experiment 2 to 4. Therefore, the \gls{rrt}* algorithm produced the shortest routes in all experiments.

Although the proposed approach did not produce the shortest routes, it produced the safest routes in all four experiments. The \gls{nc} column in Table~\ref{tab:results} shows the number of collisions or crashes for the proposed, \gls{pso} based, greedy and \gls{ga} based, \gls{rrt}, and \gls{rrt}* algorithms. The total number of crashes in all experiments was 0, 14, 9, 18, and 29, respectively. As stated in Section~\ref{sec:system}, the proposed system provides a safety-first approach in which a hazard-free, safe operation of the \gls{uav} swarm takes precedence over any other objectives including the route length. This safety-first aspect of the proposed approach is evident in all experiments. The proposed approach generated slightly longer routes by trading route length for \gls{uav} safety. As a result, all \gls{uav}-to-\gls{uav}, \gls{uav}-to-static-obstacle, and \gls{uav}-to-moving-obstacle collisions were avoided and all drones successfully completed their maneuvers.

The last column in Table~\ref{tab:results} shows the \gls{t} results of the five algorithms in milliseconds. The results show that the \gls{rrt} algorithm took the least amount of time to run and produce the results in all experiments, while the proposed approach performed second fastest in Experiment 2 and 3, and the greedy and \gls{ga} based algorithm performed second fastest in Experiment 1 and 4. Therefore, the results show that the proposed algorithm has a low computation overhead and it generates safe and efficient routes in a reasonable amount of time.

Experiment 1 results show that the proposed system is suitable for smaller problem instances. The performance and scalability of the proposed system are further demonstrated in the results of Experiment 2 which produced drone routes for a larger problem instance. In Experiment 3, the drones encountered more obstacles on their ways because the flying zone was cluttered with static and moving obstacles. It forced them to take longer routes to their destinations, but the proposed approach managed to successfully avoid all obstacles and collisions and routed all drones to their destinations. It shows that the proposed system is also suitable for densely populated, cluttered flying zones. Finally, Experiment 4 results show that the proposed approach is also suitable for complex problem instances involving high risks of drone collisions.


\begin{table}[!t]
\caption{Comparison of the results with a \glsentrytext{pso} based approach~\cite{Sujit:2009}, a greedy heuristics and \glsentrytext{ga} based approach~\cite{Silva:Arantes:2017}, and two sampling-based path planning algorithms called \glsentrytext{rrt}~\cite{Lavalle:98} and \glsentrytext{rrt}*~\cite{Karaman:2011}}
\label{tab:results}
\centering
\begin{tabular}{|l|c|c|c|c|c|} 
\hline
\textbf{Approach} & \textbf{Experiment} & \textbf{\glsentrytext{arl}} & \textbf{\glsentrytext{llr}} & \textbf{\glsentrytext{nc}} & \textbf{\glsentrytext{t}(ms)} \\
\hline

\multirow{4}{*}{Proposed}
& 1 & 17 & 36 & \textbf{0} & 670 \\ \cline{2-6}
& 2 & 34 & 62 & \textbf{0} & 683 \\ \cline{2-6}
& 3 & 20 & 47 & \textbf{0} & 642 \\ \cline{2-6}
& 4 & 36 & 97 & \textbf{0} & 637 \\ \hline

\multirow{4}{*}{PSO Based}
& 1 & 25 & 49 & 3 & 949 \\ \cline{2-6}
& 2 & 54 & 71 & 1 & 894 \\ \cline{2-6}
& 3 & 35 & 53 & 5 & 932 \\ \cline{2-6}
& 4 & 34 & 91 & 5 & 793 \\ \hline

\multirow{4}{*}{Greedy and GA}
& 1 & 26 & 47 & 1 & 600 \\ \cline{2-6}
& 2 & 53 & 69 & \textbf{0} & 715 \\ \cline{2-6}
& 3 & 36 & 56 & 4 & 714 \\ \cline{2-6}
& 4 & 34 & 94 & 4 & 627 \\ \hline

\multirow{4}{*}{\Gls{rrt}}
& 1 & 16 & 36 & 2 & \textbf{481} \\ \cline{2-6}
& 2 & 29 & \textbf{58} & 4 & \textbf{598} \\ \cline{2-6}
& 3 & 20 & 44 & 7 & \textbf{610} \\ \cline{2-6}
& 4 & 34 & \textbf{86} & 5 & \textbf{599} \\ \hline

\multirow{4}{*}{\Gls{rrt}*}
& 1 & \textbf{15} & \textbf{29} & 3 & 612 \\ \cline{2-6}
& 2 & \textbf{27} & \textbf{58} & 4 & 721 \\ \cline{2-6}
& 3 & \textbf{18} & \textbf{37} & 9 & 738 \\ \cline{2-6}
& 4 & \textbf{30} & 89 & 13 & 688 \\ \hline

\end{tabular}
\end{table}

\section{Related Work}
\label{sec:related_work}

The problem of motion safety of semi-autonomous robotic systems is currently attracting significant research attention. A comprehensive overview of the problems associated with autonomous mobile robots is given in~\cite{siegwart:2011}. The analysis carried out in~\cite{Fraichard:2007} shows that the most prominent routing schemes do not guarantee motion safety. Our approach resolves this issue and ensures not only safety but also provides efficient, online routing.

Macek et al.~\cite{Macek:2008} proposed a layered architectural solution for robot navigation. They focused on the problem of safe navigation of a vehicle in an urban environment. They also distinguished between global route planning and collision avoidance control. However, in their work, they focused on the safety issues associated with the navigation of a single vehicle and did not consider the problem of collision-free path generation and navigation in the context of fleets or swarms of robots. Aniculaesei et al.~\cite{Aniculaesei:2016} presented a formal approach that employs formal verification to ensure motion safety. They used UPPAAL model checker\footnote{\url{http://www.uppaal.org/}} to verify that a moving robot engages brakes and safely stops upon detection of an obstacle. Since our proposed system does not assume any a priori knowledge on the numbers and locations of the static and moving obstacles and does not depend on a preliminary, off-line motion planning phase, the safety requirements can not be verified before the start of the mission. Therefore, we did not employ formal verification. The solution proposed in our work is fast and flexible as it dynamically generates and recomputes the drone routes in an online manner and avoids unnecessary stopping of the drones.

Petti and Fraichard~\cite{Petti:2005} proposed an approach that relies on partial motion planning to ensure safety. They state that calculation of an entire route is such a complex and compute-intensive problem that the only viable solution is a computation of the next safe states and navigation within them. Their solution supports navigation of a single vehicle. In our work, we have discretized the flying zone and have developed a highly efficient system that computes the next safe states for an entire swarm and provides a mechanism for online path generation and collision avoidance.

A comprehensive literature review on motion planning algorithms for \glspl{uav} can be found in~\cite{Goerzen:2009}. The approaches reviewed in~\cite{Goerzen:2009} are applicable to a preliminary, off-line motion planning phase to plan and produce an efficient path or trajectory for a \gls{uav} before the start of the mission. Our proposed system does not depend on a planning phase and produces efficient, collision-free paths for an entire swarm in an online manner. A more recent survey on motion planning of \glspl{uav} can be found in~\cite{Kendoul:2012}.

Augugliaro et al.~\cite{Augugliaro:2012} presented an algorithm for generating collision-free trajectories for a quadrotor fleet. They focused on a planned approach that generates feasible paths ahead of time. LaValle~\cite{Lavalle:98} and Karaman and Frazzoli~\cite{Karaman:2011} presented sampling-based path planning algorithms called \gls{rrt} and \gls{rrt}*, respectively. \Gls{rrt} was designed to efficiently explore high-dimensional spaces by incrementally building a tree. \Gls{rrt}* is an extension of \gls{rrt}. It was designed to plan optimal paths.

Majd et al.~\cite{Majd:2018:CEC, Majd:2018:PDP} proposed a path planning and navigation approach for swarms of drones. They combined offline path planning with an online navigation approach and used machine learning and evolutionary algorithms to generate efficient paths while maximizing safety of the drones in the swarm. They also used collision prediction and drone reflexes to prevent collisions with unforeseen obstacles. In comparison, this paper presents an online, collision-free path generation and navigation approach, which does not need offline path planning.

Dong et al.~\cite{Dong:2009} presented a software platform for online cooperative control of multiple \glspl{uav}. Their work focuses on monitoring and control of multiple \glspl{uav} from a ground control station. The approach does not generate paths for the \glspl{uav}. Instead, the complete flight information (including the \glspl{uav} paths) are provided to the ground control station that sends control commands to the \gls{uav} fleet. Olivieri~\cite{Olivieri:2015} and Olivieri and Endler~\cite{Olivieri:2015a} presented an approach for movement coordination of swarms of drones using smart phones and mobile communication networks. They used \gls{cep}, but only to analyze and evaluate the formation accuracy of the swarm. Moreover, their work focuses on the internal communication of the swarm and does not provide a solution for collision-free path generation. B{\"u}rkle et al.\cite{Burkle:2011} proposed a multi-agent system architecture for team collaboration in a swarm of drones. They also developed a simulation platform for patrolling or surveillance drones which monitor a protected area against potential intrusions. However, they did not address path planning and collision avoidance for the swarm.

Ivanovas et al.~\cite{Ivanovas:2018} proposed an obstacle detection and avoidance approach for a \gls{uav}. Their approach uses computer vision techniques for detecting static obstacles in stereo camera images. The main focus of their approach is on how some block matching algorithms can be used for obstacle detection. They did not present a path planning and collision avoidance approach for multiple \glspl{uav}. Barry and Tedrake~\cite{Barry:2015} proposed an obstacle detection algorithm for \glspl{uav} that allows to detect and avoid collisions in an online manner. Similarly, Lin~\cite{Lin:2015} presented an online path planner for \glspl{uav} that detects and avoids moving obstacles. These approaches are only applicable for individual \glspl{uav} and do not provide support for a swarm of \glspl{uav}. In our work, we assumed that each \gls{uav} is equipped with an adequate obstacle sensing and detection capability and does not require any additional support for obstacle detection. Therefore, we focused on collision prediction and avoidance and online path generation and navigation for swarms of \glspl{uav}.

Sujit and Beard~\cite{Sujit:2009} proposed a \gls{pso} based path planning algorithm for swarms of drones. In their approach, whenever a drone detects a moving obstacle, the \gls{pso} based algorithm generates a new path for the drone depending on the time allowed to compute a new path before the collision can occur. Silva Arantes et al.~\cite{Silva:Arantes:2017} presented a \gls{uav} path planning approach for critical situations requiring an emergency landing of the \gls{uav}. Their approach uses greedy heuristics and \glspl{ga} to generate and optimize feasible paths under different types of critical situations caused by equipment failures.

In Section~\ref{sec:evaluation}, we have presented a comparison of the results of our proposed approach with Sujit and Beard~\cite{Sujit:2009}'s \gls{pso} based approach, Silva Arantes et al.~\cite{Silva:Arantes:2017}'s greedy heuristics and \gls{ga} based approach, LaValle~\cite{Lavalle:98}'s \gls{rrt} algorithm, and Karaman and Frazzoli~\cite{Karaman:2011}'s \gls{rrt}* algorithm. The results show that our proposed approach produced the safest routes in all four experiments. Therefore, the proposed approach outperformed the \gls{pso} based, greedy heuristics and \gls{ga} based, \gls{rrt}, and \gls{rrt}* approaches with respect to drone safety.

\glsresetall
\section{Conclusions} 
\label{sec:conclusion}

In this paper, we presented an online, collision-free path generation and navigation system for swarms of \glspl{uav}. The proposed system uses geographical locations of the \glspl{uav} and of the successfully detected, static and dynamically appearing, moving obstacles to predict and avoid: (1) \gls{uav}-to-\gls{uav} collisions, (2) \gls{uav}-to-static-obstacle collisions, and (3) \gls{uav}-to-moving-obstacle collisions. It comprises three main components: (1) a \gls{cep} and collision prediction module, (2) a mutually-exclusive locking mechanism, and (3) a collision avoidance mechanism. The \gls{cep} and collision prediction module leverages efficient runtime monitoring and \gls{cep} to make timely predictions. The mutually-exclusive locking mechanism prevents multiple \glspl{uav} from attempting to fly to the same location at the same time. The collision avoidance mechanism tries to find best ways to prevent the \glspl{uav} from colliding into one another and with the successfully detected static and moving obstacles in the flying zone. Therefore, a distinctive feature of the proposed system is its ability to foresee risks of collisions in an online manner and proactively find best ways to avoid the predicted collisions in order to ensure safety of the entire swarm.

We also presented a simulation-based implementation of the proposed system along with an experimental evaluation involving a series of experiments and compared our results with the results of four existing approaches. The results showed that the proposed system successfully predicts and avoids all three kinds of collisions in an online manner. Moreover, it generates safe and efficient \gls{uav} routes, efficiently scales to large-sized problem instances involving dozens of \glspl{uav} and obstacles, and is suitable for densely populated, cluttered flying zones and for scenarios involving high risks of \gls{uav} collisions.

As part of our future work, we plan to implement the proposed system in a more realistic simulation environment that allows to take into account complex physical phenomena and uncontrolled environment variables. Moreover, we want to test and evaluate our system for heterogeneous drones that may have diverse capabilities and fly at different speeds. Finally, an adequate support and online mechanisms to handle and control the situations arising from imprecise information of \gls{uav} locations and internal drone failures during mission execution are also planned as future works.

\section*{Acknowledgments}

The work was supported by the Academy of Finland projects OpenCPS: Open Integrated Framework for Accelerating Development of Resilient CPS and CoRA: Continuous Resilience Assurance of Complex Software-Intensive Systems.

\bibliographystyle{abbrv} 
\bibliography{bibliography}

\end{document}